\PassOptionsToPackage{unicode}{hyperref}
\PassOptionsToPackage{hyphens}{url}
\documentclass[11pt]{article}

\usepackage[T1]{fontenc}
\usepackage[utf8]{inputenc}
\usepackage{lmodern}
\usepackage{textcomp}
\usepackage{amsmath,amssymb}
\usepackage{xcolor}
\usepackage{graphicx}
\usepackage{longtable,booktabs,array}
\usepackage{calc}
\usepackage{etoolbox}
\usepackage{fancyvrb}
\usepackage[margin=1in]{geometry}
\usepackage{parskip}
\usepackage{caption}
\usepackage{float}

\usepackage{newunicodechar}
\newunicodechar{ʿ}{\textquoteleft}
\newunicodechar{ʾ}{\textquoteright}
\newunicodechar{ḥ}{\d{h}}  \newunicodechar{Ḥ}{\d{H}}
\newunicodechar{ḍ}{\d{d}}  \newunicodechar{Ḍ}{\d{D}}
\newunicodechar{ṣ}{\d{s}}  \newunicodechar{Ṣ}{\d{S}}
\newunicodechar{ṭ}{\d{t}}  \newunicodechar{Ṭ}{\d{T}}
\newunicodechar{ẓ}{\d{z}}  \newunicodechar{Ẓ}{\d{Z}}
\newunicodechar{ā}{\={a}}  \newunicodechar{Ā}{\={A}}
\newunicodechar{ī}{\={\i}} \newunicodechar{ū}{\={u}}
\newunicodechar{→}{$\rightarrow$}
\newunicodechar{×}{$\times$}
\newunicodechar{≥}{$\geq$}
\newunicodechar{≤}{$\leq$}
\newunicodechar{≈}{$\approx$}
\newunicodechar{−}{$-$}
\newunicodechar{λ}{$\lambda$}
\newunicodechar{τ}{$\tau$}
\newunicodechar{μ}{$\mu$}

\usepackage{bookmark}
\IfFileExists{xurl.sty}{\usepackage{xurl}}{}
\hypersetup{hidelinks, pdfcreator={LaTeX}}

\providecommand{\tightlist}{\setlength{\itemsep}{0pt}\setlength{\parskip}{0pt}}
\title{\bfseries Grading the Narrators: An Isnād--Rijāl Framework for\\
Claim-Level Provenance in Multi-Agent Knowledge Systems}
\author{Ali Zahid Raja\\
\normalsize Independent Researcher \quad \texttt{alizahidrajaa@gmail.com} \quad ORCID: 0009-0003-7875-4590}
\date{27 July 2026}

\begin{document}
\maketitle

\begin{abstract}

Modern multi-agent knowledge systems increasingly accumulate knowledge through
chains of autonomous transformations rather than direct retrieval. Existing
provenance work records \emph{what happened} --- execution traces, tool calls,
evidence links --- and source-reliability estimation is long established
(truth discovery, reputation systems). What is missing is an operational
framework that attaches graded, per-domain transmitter reliability to
claim-level transmission chains, with completeness semantics,
transformation-typed aggregation, decoupled content criticism, and
serve/review/quarantine routing for compiled multi-agent knowledge.

Classical Islamic hadith science confronted a structurally similar problem:
deciding whether knowledge transmitted through chains of human narrators should
be accepted. Over centuries it developed a rigorous methodology --- isnād (a
complete transmission chain attached to every claim), rijāl (systematic grading
of each narrator's integrity and precision), weakest-link chain evaluation,
corroboration through independent chains, and matn criticism (content evaluated
independently of chain quality). This paper transfers that methodology to AI
system design.

The framework contributes: (1) a formal mapping from hadith-science concepts to
multi-agent knowledge pipelines; (2) a relational schema implementing claim
chains and a graded narrator registry; (3) a decision matrix combining chain
grade with content criticism to yield concrete serve/review/quarantine actions;
and (4) an empirical evaluation on 20{,}000 claims extracted from real
undergraduate physics textbooks.

The evaluation validates two mechanisms and is explicit about a third. Weakest-link
grading correctly quarantined every claim whose chain contained a rejected
narrator, with each quarantine traceable to the grade that caused it. The
jarḥ--taʿdīl loop independently recovered three of four narrator grades from
audit evidence --- and missed the fourth, which carried the highest fault rate
in the experiment, because it was too rare in the calibration split to be
graded at all. Independent-chain corroboration fires across three corpora of
increasing difficulty, with negative controls on the Wikipedia corpus. Two
results are reported as inconclusive rather than positive: a matched-coverage
comparison against the baseline could not be completed, because the framework
could not be driven above 4.8\% coverage with the reference content critic;
and the confidence-gated baseline is uninformative by construction and
therefore supports no claim about real model self-confidence.

The framework targets factual knowledge accumulation, not open-ended or creative
generation, and is intentionally modular: different deployments may instantiate
different grading functions while preserving the same architectural principles.
This is a systems-architecture and epistemology paper. It proposes a framework,
a schema, and an evaluation, and it is explicit throughout about which claims
the evidence does and does not yet support.


\bigskip\hrule\bigskip

\end{abstract}
\subsection*{1. Introduction}

Modern AI systems increasingly communicate not raw evidence but transformed
knowledge. A single factual claim may pass through document retrieval,
extraction, summarization, knowledge integration, and answer synthesis before
reaching a user. While provenance systems can often reconstruct this chain after
the fact, they provide limited guidance for answering the question users
ultimately care about: why should this particular chain of transformations be
trusted?

Consider the pipeline behind a single answered question in a modern
LLM-maintained knowledge base: a scraper extracts text from a source; an
ingestion agent analyzes it, reconciles it against existing pages, and writes a
compiled article; an answer agent later reads several such articles and
synthesizes a cited response. The claim the user receives has passed through at
least four hands --- the original source, the extraction step, the ingestion
model, the synthesis model --- and each hand can drop, distort, or invent.
Current systems attach page-level citations (``this answer drew on these
documents''), and a growing body of provenance work records execution traces of
what each agent did [Souza et al.\ 2025; arXiv:2606.04990]. But neither answers
the question a careful reader actually asks: \emph{this specific claim, having
passed through these specific transformers --- should I trust it?}

This problem is not new. Its most rigorous pre-modern solution took shape over
roughly twelve centuries, beginning nearly fourteen hundred years ago.

After the death of the Prophet Muhammad (632 CE), Muslim scholars faced an
epistemic crisis of exactly this shape: statements attributed to him circulated
through lengthening chains of human transmitters, with fabrication, error, and
honest distortion all in play. The response was not to trust the content on its
plausibility, nor to trust chains on their length, but to build a formal
discipline --- ʿilm al-ḥadīth --- whose core insight was that \textbf{the
trustworthiness of a claim is a function of the graded reliability of every
individual who transmitted it}. Every accepted narration carries its isnād: the
complete, gap-free chain of narrators. A parallel discipline, ʿilm al-rijāl
(``the science of the transmitters''), maintained biographical assessments of
each narrator's integrity (ʿadālah) and precision (ḍabṭ), continuously updated
through a formal criticism-and-accreditation process (al-jarḥ wa-l-taʿdīl). A
chain is graded by its weakest link; independent corroborating chains can
upgrade a claim; and even a flawless chain does not exempt the content itself
from criticism (naqd al-matn), because a sound chain can carry a subtly
defective text (ʿilal) [Brown 2017; Ibn al-Ṣalāḥ, 13th c.].

Substitute ``narrator'' with ``agent, model version, or scraper'' and this is a
design specification for claim-level trust in multi-agent AI systems --- one
that existing provenance frameworks, which record chains but do not grade the
transmitters, have not articulated. I call the resulting framework
\textbf{ISNAD}; where the lowercase \emph{isnād} appears, it denotes the
classical concept rather than the system.

\textbf{Contributions.}
(1) A formal mapping from isnād--rijāl methodology to multi-agent knowledge
pipelines (§4).
(2) A relational schema --- claim chains plus a graded narrator registry ---
implementing the mapping, with an open-source reference implementation (§5).
(3) A decision matrix combining chain grade with content criticism that yields
concrete serve/review/quarantine actions (§4.4).
(4) A case study in which a prototype self-maintaining knowledge base surfaced
19 genuine cross-framework contradictions in undergraduate physics texts,
demonstrating the matn-criticism substrate (§6).
(5) An evaluation on 20{,}000 claims from real physics textbooks that validates
weakest-link quarantine and corroboration, reports a partial failure of the
jarḥ--taʿdīl loop, and reports two analyses as inconclusive (§8).
(6) An explicit account of what remains unvalidated and what would validate it
(§7, §8.7).

\textbf{Direction of transfer.} A substantial literature applies AI \emph{to}
hadith texts --- machine learning for chain analysis, authenticity
classification, narrator-network mining [Int'l J. Noesantara Islamic Studies
2025; Digital Muslim Review 2025]. This paper inverts the direction: it applies
hadith \emph{methodology to} AI system design. To the best of my knowledge, no
prior work systematically adapts the isnād--rijāl methodology into an
operational provenance architecture for multi-agent knowledge systems.

\bigskip\hrule\bigskip

\subsection*{2. Background}

This section reviews the two bodies of knowledge that anchor the framework: the
classical hadith transmission science, and the current state of provenance and
trust in multi-agent AI systems. It then traces a deeper lineage across
centuries of convergent solution design.

\subsubsection*{2.1 Source reliability across traditions: the deep lineage}

The problem of assessing transmitted information --- determining which claims to
trust when they arrive through chains of human or machine intermediaries --- has
been independently solved, in partial forms, across multiple intellectual
traditions. This section traces that lineage to establish that ISNAD's
contribution is not novelty for its own sake, but the transfer of the most
operationally complete solution to a domain that lacks it.

\textbf{Wave 1 --- Pre-computational foundations.} Condorcet's jury theorem
(1785) demonstrated that independent witnesses each better than random produce
error that falls with agreement --- the mathematical ancestor of
corroboration-based error reduction. If each voter has probability $p > 0.5$ of
being correct, a majority vote approaches certainty as the group grows. This is
independence-based error reduction, the formal analogue of mutābaʿāt. Hume
(1748), in ``Of Miracles,'' argued that testimony credibility must be weighed
against witness reliability, introducing the intuition that transmitters
themselves must be graded. Anglo-American evidence law codified these
intuitions: the hearsay-within-hearsay rule (U.S. FRE 805) requires each link in
a multi-hop testimony chain to independently qualify --- a legal weakest-link
principle. Chain-of-custody requirements are the legal analogue of ittiṣāl
(unbroken transmission).

\textbf{Wave 2 --- Statistical and trust-theoretic formalization.} Dempster
(1967) introduced upper and lower probabilities, laying groundwork for evidence
combination with source-reliability discounting; Shafer (1976) developed this
into a full mathematical theory of evidence. Dawid and Skene (1979) provided the
statistical skeleton for grading annotators: maximum-likelihood estimation of
observer error rates via EM, even when true labels are unknown. Jøsang (2001)
introduced subjective logic --- trust discounting along chains plus consensus
fusion, the closest formal ancestor of weakest-link aggregation with
corroboration. This was extended through beta reputation (Jøsang \& Ismail,
2002), EigenTrust (Kamvar et al., 2003), ReGreT (Sabater \& Sierra, 2001), and
FIRE (Huynh et al., 2006).

\textbf{Wave 3 --- Truth discovery and knowledge fusion at scale.} Yin, Han and
Yu (2008) introduced TruthFinder, jointly estimating truth and source
reliability from conflicting web data. Pasternack and Roth (2010, 2013) extended
this with prior-knowledge integration and latent credibility analysis. Google's
Knowledge Vault (Dong et al., 2014) fused web extractions into a probabilistic
knowledge base; Knowledge-Based Trust (Dong et al., 2015) estimated web source
trustworthiness from factual correctness. Database provenance (Cui \& Widom,
2000; Buneman et al., 2001) and Semantic Web provenance (Gil \& Artz, 2007;
Hartig \& Zhao, 2009) addressed lineage tracking.

\textbf{Positioning --- what each family lacks.} Truth discovery grades sources
but treats the path from source to serving as transparent: no transmission
chains, no transformation types, no completeness semantics. Subjective logic
supplies chain-discounting arithmetic but no operational registry lifecycle
(per-domain grades, version-bump resets, quarantine), no decoupled content
criticism, and no serve/review/quarantine routing. Multi-agent reputation
systems grade peer agents for partner selection, not claim-level provenance in
compiled knowledge. ISNAD's contribution is the assembled operational protocol
--- chains, a graded per-domain registry, transformation-typed weakest-link
aggregation, gated corroboration, dual criticism, and a decision matrix ---
transferred intact from the tradition that ran it as a complete system.

\subsubsection*{2.2 A primer on hadith transmission science}

The following concepts, formalized between roughly the 8th and 13th centuries CE
and systematized in works such as Ibn al-Ṣalāḥ's \emph{Muqaddimah} and
al-Dhahabī's narrator-criticism compendium \emph{Mīzān al-Iʿtidāl}, constitute
the framework this paper adapts. Jonathan Brown's \emph{Hadith} [2017] is the
standard English academic treatment.

\begin{table}[h]
\centering\small
\begin{tabular}{@{}p{0.17\linewidth}p{0.40\linewidth}p{0.35\linewidth}@{}}
\toprule
\textbf{Term} & \textbf{Meaning} & \textbf{This paper's usage} \\
\midrule
isnād & The complete chain of transmitters attached to a narration & The agent chain attached to a claim \\
matn & The transmitted content itself & The claim text \\
rijāl & ``The transmitters''; the discipline of evaluating them & The narrator registry \\
ʿadālah & A narrator's integrity/uprightness & Resistance to manipulation; source trustworthiness \\
ḍabṭ & A narrator's precision and retention accuracy & Measured error rate of an agent/model \\
al-jarḥ wa-l-taʿdīl & Formal criticism and accreditation of narrators & The registry update process (evals, audits) \\
ittiṣāl / munqaṭiʿ & Chain continuity / a broken chain & Complete trace / missing trace segment \\
ṣaḥīḥ, ḥasan, ḍaʿīf, mawḍūʿ & Sound, good, weak, fabricated (grades) & Trust tiers for claims \\
mutābaʿāt & Independent corroborating chains & Multiple disjoint agent chains asserting one claim \\
ʿilal & Hidden defects: sound-looking chain, defective content & Chain passes but content criticism fails \\
muḥaddith & The expert scholar who adjudicates difficult cases & The human reviewer in the loop \\
\bottomrule
\end{tabular}
\caption{Correspondence between hadith-science concepts and their usage in this framework.}
\label{tab:primer}
\end{table}

Five structural principles matter for the transfer:

\begin{enumerate}
\tightlist
\item \textbf{Every claim carries its full chain.} A narration without a complete
  isnād is inadmissible by default; a chain with a gap (munqaṭiʿ) is
  automatically downgraded regardless of content quality.
\item \textbf{Transmitters are graded individually and continuously.} Rijāl
  scholarship maintained living biographical databases; a narrator's grade could
  fall when errors surfaced (jarḥ) or rise with corroborated accuracy (taʿdīl).
\item \textbf{The chain is graded by its weakest link.} One weak narrator caps
  the grade of the whole chain, however strong the others.
\item \textbf{Corroboration can upgrade.} A claim carried by multiple
  \emph{independent} chains (mutābaʿāt) can rise a grade --- the epistemics of
  redundancy, formalized.
\item \textbf{Content is criticized independently of the chain.} A perfect isnād
  does not immunize the matn; contradictions with established knowledge trigger
  the ʿilal category --- hidden defect --- and expert adjudication.
\end{enumerate}

\textbf{Acknowledging the critical tradition.} The Western critical tradition on
hadith --- from Schacht's skepticism about early Islamic legal transmission
(Schacht, 1950) to Juynboll's form-critical analyses (Juynboll, 1983) --- has
questioned traditional authentication methods. Yet even within this critical
tradition the methodological architecture has analytic value. Motzki's
isnād-cum-matn analysis systematically compares transmission chains and text
variants across narrations to date and verify hadith --- a secular academic
method that uses joint chain-and-content analysis, precisely the dual-criticism
architecture this paper borrows. More recently, computational hadith studies
have applied AI to narrator disambiguation: Mosa (2025) proposed a knowledge
graph--transformer framework for narrator resolution in hadith networks. This
paper inverts that direction: rather than applying AI to hadith, it extracts
hadith's operational protocol and applies it to AI systems. The paper borrows
the methodological architecture and takes no position on theological
authentication.

\subsubsection*{2.3 Multi-agent knowledge systems and the provenance state of the art}

The system class motivating this work is the LLM-maintained, compiled knowledge
base: rather than re-deriving answers from raw documents per query
(retrieval-augmented generation), an LLM continuously compiles sources into a
structured, interlinked wiki, resolves contradictions once, and serves cited
answers from the compiled artifact --- a pattern recently popularized by
Karpathy's ``LLM Wiki'' sketch [Karpathy 2026] and related to persistent
agent-memory systems [Chhikara et al.\ 2025]. In such systems every compiled
claim is the product of a multi-step agent pipeline, which makes claim-level
trust both more important and more tractable: the transformation points are
known and instrumentable.

Provenance research for LLM agents has advanced rapidly. PROV-AGENT extends the
W3C PROV model [Moreau et al.\ 2013] to capture agent interactions in agentic
workflows [Souza et al.\ 2025]; Agent-Sentry bounds agents via execution
provenance [Sequeira et al.\ 2026]; and a recent survey organizes the field
around \emph{evidence tracing} and \emph{execution provenance}
[arXiv:2606.04990]. In parallel, claim-level factuality work --- FActScore's
atomic-claim decomposition [Min et al.\ 2023], the FEVER and VitaminC
verification benchmarks [Thorne et al.\ 2018; Schuster et al.\ 2021] --- grades
claims against evidence at a point in time. Knowledge-conflict studies show that
LLMs handle contradictory evidence poorly: they struggle to represent, surface,
or reconcile competing claims [Su et al.\ 2024; Hou et al.\ 2024; Wu et al.\
2024], and multi-agent failure taxonomies attribute many system failures to weak
verification between agents [Cemri et al.\ 2025]. ContraCrow demonstrates
LLM-driven contradiction detection against scientific literature at scale
[arXiv:2409.13740].

A third, adjacent stream applies computation \emph{to} hadith grading itself:
classifiers that sort narrations into ṣaḥīḥ/ḥasan/ḍaʿīf/mawḍūʿ from chain and
text features [Ramzy et al.\ 2023], narrator-name extraction and chain datasets
[Mghari et al.\ 2022], and --- most relevant here --- Hawramani's
\emph{HadithRank}, which models an isnād's topology using information theory,
treating independent transmission chains as redundant channels whose combined
error probability falls as $p^n$ with $n$ independent chains [Hawramani].
HadithRank is notable because it independently formalizes, inside the hadith
domain, the same epistemic principle this paper's corroboration rule (§4.3)
relies on: independent chains multiply reliability. That a mathematician grading
hadith and this paper grading agents converge on the same redundancy rule is
evidence the mapping is structural, not decorative. The direction, however, is
opposite: that stream brings computation \emph{to} hadith; this framework brings
hadith methodology \emph{to} computational systems.

A parallel body of work studies computational trust itself rather than
provenance. Subjective-logic trust models, reputation systems for multi-agent
environments, truth-discovery algorithms, and recent credibility-scoring methods
for LLM-based agents estimate the reliability of entities, sources, or claims
through probabilistic updating, consensus, or historical behavior. These
approaches address important aspects of trust calibration, but they generally do
not couple trust estimation to explicit claim-level transmission chains, graded
transformer registries, and an interaction between transmission quality and
independent content criticism. Rather than replacing these approaches, the
framework proposed here is complementary: it specifies an architectural protocol
for how trust should propagate through multi-agent knowledge transmission.
Recent work has continued this trajectory. Ebrahimi et al.\ (2025) introduced
credibility scoring for adversary-resistant multi-agent LLM systems, learning
agent trustworthiness from past contributions. RAPS (2026) proposed a
reputation-aware publish--subscribe framework with Bayesian reputation overlays
for LLM coordination. TrustTrade (2026) replaces ``uniform trust'' --- the
observed tendency of LLMs to treat heterogeneous sources identically --- with
cross-agent consistency weighting. Prakash (2026) identified the ``Provenance
Paradox'': when delegates can inflate self-reported quality, quality-based
routing can perform worse than random.

\textbf{The gap.} The provenance and factuality lines record chains, verify
claims against evidence, and detect conflicts, but none maintains a \emph{graded
registry of the transmitters themselves} with a defined update process,
aggregates chain trust by a weakest-link rule with a corroboration escape hatch,
or defines the interaction between chain quality and content criticism. The
AI-for-hadith line has formalized narrator grading and chain redundancy, but for
authenticating historical texts, not for governing live multi-agent knowledge
pipelines. Combining a graded transmitter registry, weakest-link-plus-corroboration
aggregation, chain completeness as an epistemic property, and a chain-quality
$\times$ content-criticism decision procedure --- as an operational framework
for AI systems --- is the contribution rijāl methodology makes here.

\bigskip\hrule\bigskip

\subsection*{3. Threat model and scope}

The framework targets \emph{epistemic} degradation in cooperative pipelines:
extraction errors, model hallucination during compilation, stale or unreliable
sources, silent quality drift across model versions, and unresolved
contradictions between sources. It also covers one adversarial case natively: a
source that attempts prompt injection is, in rijāl terms, a narrator whose
ʿadālah is compromised --- and memory/knowledge-base poisoning attacks against
agents are an active threat [Chen et al.\ 2024].

Out of scope: Byzantine collusion among agents (cf. Lamport, Shostak \& Pease,
1982), cryptographic attestation of traces (complementary; see [Sequeira et al.\
2026]), and fairness questions raised by grading \emph{human} contributors,
which I flag in §7 rather than resolve. Also out of scope by construction is
open-ended or creative generation: ISNAD grades claims that can, in principle,
be true or false and checked against a corpus. It is a framework for knowledge
accumulation, not for tasks where there is no fact of the matter to transmit.
A further limitation of scope, discussed in §7, is that the framework as
specified assumes claims with determinate truth values; claims that are
inherently probabilistic or provisional require an extension not developed here.

\bigskip\hrule\bigskip

\subsection*{4. The Isnād--Rijāl Framework}

\subsubsection*{4.0 Design principles}

Before presenting the framework's mechanisms, it is useful to state the
principles that govern every design choice. These are not additional components;
they summarize the epistemic commitments inherited from the isnād--rijāl
methodology and guide the architecture that follows.

\begin{enumerate}
\tightlist
\item \textbf{Every claim carries its transmission chain.} Every factual
  assertion must retain an explicit record of the transformations through which
  it reached the knowledge base.
\item \textbf{Every transmitter maintains an evolving reliability record.}
  Sources, agents, models, and human contributors accumulate evidence over time,
  allowing their reliability to improve or degrade as new observations arrive.
\item \textbf{Chain trust is bounded by its weakest verified transformation.}
  Trust cannot exceed the least reliable verified step in a chain, except where
  justified by independent corroboration.
\item \textbf{Independent corroboration increases justified confidence.}
  Multiple genuinely independent chains strengthen confidence in a claim without
  eliminating the need for critical evaluation.
\item \textbf{Transmission quality and content quality are evaluated
  independently.} A trustworthy chain does not guarantee a correct claim, and a
  correct claim does not compensate for an unreliable transmission history.
\end{enumerate}

\subsubsection*{4.1 Narrators and chains}

A \textbf{narrator} is any entity that originates, transforms, or transmits a
claim: an external source, a scraper version, an ingestion model at a specific
version and prompt revision, an answer model, or a human contributor. An
\textbf{isnād} for claim $c$ is the ordered, gap-free sequence of narrators from
origin to the point of serving, each link carrying timestamp, version
identifiers, and trace references.

Two rules follow directly from classical practice.

\textbf{Completeness (ittiṣāl).} A claim whose chain has a missing segment --- a
transformation not captured in traces --- is munqaṭiʿ and automatically capped
at the weak tier, regardless of how reliable the recorded narrators are. This
makes trace completeness a first-class epistemic property rather than an
operational nicety.

\textbf{Weakest link.} A single chain's grade is the minimum grade over its
narrators. A frontier synthesis model at the end of a chain cannot repair a
compromised extraction at the start of it --- a property that current
confidence-scoring practice, which often reflects only the final model's
self-assessment, does not have.

This rule needs one refinement to hold for AI pipelines rather than human ones.
In hadith transmission, a weak link is unconditionally corrupting: a narrator
with poor memory irreversibly degrades the text passing through them. AI
transformations are not all alike. A \emph{destructive} transformation ---
extraction, chunking, lossy summarization --- can only lose or corrupt
information, and there the strict minimum applies: nothing downstream recovers
what the extractor dropped. A \emph{generative} transformation --- synthesis by
a model with broad pre-training --- can either repair upstream noise or
introduce fresh corruption by favoring its parametric prior over retrieved
evidence when the two conflict, a behavior documented in ClashEval (Wu et al.,
2024). For generative steps the rule is therefore not a blind minimum but a
\emph{bounded} one: a generative narrator can raise the floor only up to its own
grade and only when corroboration supports the repair, and it can always lower
it.

This mirrors a distinction the muḥaddithūn themselves drew between riwāya
bi-l-lafẓ (transmission of exact wording) and riwāya bi-l-maʿnā (transmission by
meaning): meaning-transmission, where the narrator reconstructs rather than
copies, was held to a different and often stricter standard precisely because
the transmitter's own competence now enters the chain. Destructive steps take
the strict minimum; generative steps take a bounded, corroboration-gated
adjustment. The weakest-link principle survives; it is refined by transformation
type.

\textbf{Repeated narrators.} One edge case deserves explicit treatment. If the
same narrator appears twice in a chain, a naive minimum returns that narrator's
grade unchanged --- yet two passes through the same lossy transformation are
strictly worse than one. A faithful implementation should treat repeated
traversal of a destructive narrator as compounding rather than idempotent, and
should additionally treat the repetition as evidence against independence when
the chain is later considered for corroboration (§4.3). The reference
implementation records repetition but does not yet penalize it; this is a known
gap.

\subsubsection*{4.2 The narrator registry and al-jarḥ wa-l-taʿdīl}

The \textbf{narrator registry} is the system's rijāl compendium: one row per
narrator, carrying type, version, and grade. Grades are \textbf{ordinal first,
numeric second}: narrators are placed in tiers (reliable/acceptable/weak/rejected),
with numeric error rates attached only where calibration data exists. This
ordering matters --- classical rijāl grading was categorical precisely because
false numeric precision invites misplaced confidence, and a registry populated
with uncalibrated decimals would reproduce that failure.

The registry is updated through a defined process --- the computational
jarḥ--taʿdīl loop. This is a state machine, not a formula. Each narrator occupies
an ordinal state (reliable → acceptable → weak → rejected), and transitions are
driven by named evidence types: per-narrator evaluation harnesses (extraction
fidelity for scrapers, task-specific accuracy for models), post-hoc audits of
served claims, corroboration and contradiction outcomes, and human review
verdicts. A downgrade fires when accumulated adverse evidence crosses a
threshold; an upgrade requires sustained corroborated accuracy.

The paper deliberately does \emph{not} fix the exact transition arithmetic ---
the sliding window, the update rule, the specific thresholds --- as a
first-class part of the framework. This is a considered choice, not an omission:
a protocol specifies its state space and transition triggers and leaves tuning
constants to implementations, exactly as the W3C PROV data model [Moreau et al.\
2013] defines provenance structure without dictating how systems compute trust
over it, and as transport protocols define state transitions without fixing
congestion-control constants. Committing to a specific update formula with no
deployment data to calibrate it would manufacture false precision of exactly the
kind the framework is built to avoid. §8.6 reports what happens when the
reference policy is swept, and the answer --- that the choice materially changes
both coverage and grade recovery --- is itself an argument for leaving it open.

A model-version bump resets a narrator to ungraded until re-evaluated: version
drift is treated as a new narrator, not an inherited reputation. Because grading
is domain-conditioned (below), a version bump resets grades per domain as
evaluation data arrives, rather than all at once. This mechanism has a known
blind spot: a hosted model whose weights change without a version-string change
will silently retain a grade it no longer merits. Deployments against third-party
APIs should therefore treat scheduled re-evaluation as mandatory rather than
relying on version signals alone.

ʿAdālah and ḍabṭ map onto distinguishable properties. Ḍabṭ --- precision --- is
the measurable error rate. ʿAdālah --- integrity --- covers
manipulation-resistance: whether a source domain is reputable, whether a scraper
fences untrusted content against prompt injection, whether a human contributor's
submissions have survived verification. The mapping is imperfect (§7): models are
stochastic artifacts, not moral agents. But the \emph{functional} distinction ---
``does this narrator make honest mistakes, or can it be made to lie?'' --- carries
over cleanly, and it is a distinction current provenance schemas do not draw.

\textbf{Grading is domain-conditioned, not global.} A model's reliability is not
a single number. An LLM's error rate varies sharply with the semantic domain of
the task --- a model precise on historical dates may be unreliable on
quantum-mechanical equations --- and even varies run to run under fixed
settings. A single global grade per narrator would therefore be a naive point
estimate of a distribution that does not exist as a scalar. The framework grades
each narrator \emph{per domain}: the registry key is (narrator, domain), not
narrator alone. This is not a concession forced by the peculiarities of LLMs; it
is a return to classical practice. Rijāl scholars graded transmitters
domain-specifically as a matter of course --- a narrator could be judged
reliable (thiqah) in the hadith of one teacher and weak in another's, precise in
law and unreliable in history. The discipline never assumed a transmitter was
uniformly reliable across all material, and neither should a system grading
models.

Domain-conditioned grading carries an operational cost that §8.2 makes concrete:
it multiplies the number of registry cells that must be filled, and a narrator
that is rare within a given domain may never accumulate enough evidence to be
graded at all.

\subsubsection*{4.3 Corroboration (mutābaʿāt)}

The weakest-link rule alone would be too conservative: much useful knowledge
arrives through imperfect chains. Classical methodology solves this with
corroboration: a claim carried by $k$ independent chains --- chains sharing no
narrators --- may be upgraded. Computationally: same normalized claim, disjoint
narrator sets, independent sources.

Two constraints keep this from becoming an unbounded trust-inflation loop.
First, the upgrade is \emph{capped} --- corroboration can raise a claim toward
but not past the sound tier, and it cannot manufacture trust from chains that
are all weak: some corroborating chain must itself clear a minimum grade.
Second, the exact aggregation is, like the jarḥ--taʿdīl transition function
(§4.2), a deliberately open implementation parameter. The ordinal-first
commitment governs what the system \emph{reports} (tiers, not false-precise
decimals); it does not forbid an implementation from computing tier upgrades
numerically under the hood. HadithRank's information-theoretic model ---
combined error probability falling as roughly $p^n$ across $n$ independent
chains [Hawramani] --- is precisely one such numeric instantiation, developed
independently inside the hadith domain.

The framework specifies the \emph{rule} (independent chains upgrade, capped,
gated on a minimum grade) and leaves the \emph{arithmetic} to implementations.
This gives the framework the redundancy epistemics that single-chain confidence
scores lack, and it creates the right incentive at the system level --- breadth
of independent sourcing, not repeated re-ingestion through the same pipeline.
The design parallels Condorcet's insight (1785) that independent agreement
reduces error and Shafer's evidence combination (1976), with HadithRank as the
within-domain instantiation.

The word doing the most work in this rule is \emph{independent}, and §7 returns
to how easily it can be overstated.

\subsubsection*{4.4 Dual criticism and the decision matrix}

The chain grades the \emph{transmission}; matn criticism grades the
\emph{content} --- in a compiled knowledge base, contradiction detection against
the existing corpus at ingest time. The two verdicts interact according to a
decision matrix, which is the framework's most directly actionable artifact.

\begin{table}[h]
\centering\small
\begin{tabular}{@{}p{0.19\linewidth}p{0.24\linewidth}p{0.24\linewidth}p{0.24\linewidth}@{}}
\toprule
& \textbf{Matn: consistent} & \textbf{Matn: contradiction} & \textbf{Matn: unverifiable} \\
\midrule
\textbf{Ṣaḥīḥ-tier chain} (complete; all narrators graded reliable)
& Serve directly; cache
& \textbf{ʿIlal flag} → human review. Highest-value review case.
& Serve with caveat \\
\addlinespace
\textbf{Ḥasan-tier chain} (complete; ≥1 narrator ungraded or mid-tier)
& Serve with explicit confidence caveat; seek corroboration
& Hold in review queue; do not serve
& Hold in review queue \\
\addlinespace
\textbf{Ḍaʿīf-tier chain} (weak narrator, or munqaṭiʿ)
& Hold; seek corroborating chain before serving
& Quarantine
& Hold in review queue \\
\addlinespace
\textbf{Mawḍūʿ-tier chain} (rejected narrator --- e.g.\ known injection source)
& Reject and log; quarantine the narrator
& Reject and log; quarantine the narrator
& Reject and log; quarantine the narrator \\
\bottomrule
\end{tabular}
\caption{The decision matrix. Chain grade (rows) $\times$ content-criticism verdict (columns) $\rightarrow$ action.}
\label{tab:matrix}
\end{table}

\textbf{The third column is not a formality.} Content criticism is a partial
function: a critic that cannot evaluate a claim must be able to say so rather
than defaulting to ``consistent.'' Unverifiable is the classical position of
tawaqquf --- suspension of judgment --- and routing it conservatively is what
makes the framework safe under a weak critic. It is also, as §8.6 shows, the
binding constraint on practical coverage: with a critic that returns
unverifiable on most real prose, ḥasan-tier claims cannot graduate to serving,
and coverage is capped at the review budget regardless of how well chain grading
performs. A framework that omitted this column would look far more capable on
paper and behave far worse in deployment.

Three design defaults embedded here deserve emphasis.

First, \textbf{contradictions are adjudicated by humans by default} --- the
muḥaddith role. This is not conservatism for its own sake: knowledge-conflict
research consistently finds LLMs unreliable at representing and reconciling
competing evidence [Su et al.\ 2024; Hou et al.\ 2024; Wu et al.\ 2024], and
during development of the prototype described in §6, early evaluations found
that LLM synthesis handled contradiction cases \emph{worse} than an extractive
baseline until contradiction markers were made explicitly visible in context ---
after which auto-resolution was still disabled in favor of gated review. The
classical discipline reached the same conclusion about hard cases a millennium
earlier: they go to the expert.

Second, the ṣaḥīḥ-chain-with-contradiction cell is the system's most informative
signal, not an error state --- it is where either a trusted source has changed
the world's state or the corpus contains a latent defect, and both are exactly
what a self-maintaining knowledge base exists to catch.

Third, the mawḍūʿ tier is the framework's native defense against knowledge-base
poisoning. Adversarial prompt-injection and memory-poisoning attacks against
agents are an active and demonstrated threat [Chen et al.\ 2024]: a compromised
source or a manipulated agent is, in rijāl terms, a fabricator --- a narrator
whose ʿadālah has failed. Grading narrators for integrity, not merely precision,
means the framework quarantines a source that attempts injection rather than
merely noting low confidence in its output. The rejected tier is thus an active
containment action, and it connects claim-level provenance directly to the
AI-security literature on agent poisoning.

\subsubsection*{4.5 A worked example}

Consider the claim \emph{``the momentum of a photon is $p = h/\lambda$''},
ingested from an openly licensed physics text.

\textbf{Chain:} [OpenStax University Physics Vol.\ 3 (source; publisher-trusted,
ʿadālah high)] → [pdf-scraper v1.2 (ḍabṭ graded: high extraction fidelity on
text-layer PDFs)] → [ingest-analysis model $M$@v (ungraded --- recent version
bump)] → [ingest-renderer $M$@v (ungraded)].

\textbf{Chain grade:} complete (ittiṣāl holds), but two narrators ungraded →
ḥasan-tier by the weakest-link rule.

\textbf{Matn criticism:} the corpus already contains a page asserting momentum
as $p = mv$ without regime qualification. Contradiction detected.

\textbf{Matrix cell:} ḥasan × contradiction → review queue; do not serve.

\textbf{Adjudication:} a human reviewer recognizes both claims as valid in
different regimes --- classical and quantum --- and updates both pages with
regime qualifiers, resolving the contradiction \emph{once, at ingest time},
rather than leaving every future query to renegotiate it. The resolution event
feeds back: the ingestion model's contradiction-detection behavior on this case
becomes taʿdīl evidence toward its grade; both claims' chains record the
adjudication.

This is the framework operating end to end: chain capture, weakest-link grading,
dual criticism, matrix-routed action, registry update. It is implemented as a
runnable example and an integration test in the reference implementation.

\subsubsection*{4.6 What rijāl adds over existing provenance}

Compressed to its differences from the state of the art [Souza et al.\ 2025;
arXiv:2606.04990; Moreau et al.\ 2013], the contribution is a package of five
interlocking mechanisms. Table~\ref{tab:ancestors} anchors each to its strongest
ancestor.

\begin{table}[h]
\centering\small
\begin{tabular}{@{}p{0.20\linewidth}p{0.32\linewidth}p{0.40\linewidth}@{}}
\toprule
\textbf{Mechanism} & \textbf{Strongest ancestor} & \textbf{What ISNAD adds} \\
\midrule
Chain aggregation & Jøsang subjective logic (2001) --- trust discounting along chains & Transformation typing (riwāya bi-l-lafẓ vs.\ bi-l-maʿnā); completeness semantics \\
\addlinespace
Source registry & Truth discovery (Yin et al., 2008; Dong et al., 2014) --- source reliability estimation & Per-domain grading; version-bump resets; quarantine lifecycle \\
\addlinespace
Content criticism & FEVER-style fact-checking & Decoupled from chain quality; operates on matn independently; explicit unverifiable verdict \\
\addlinespace
Agent grading & MAS reputation (EigenTrust, ReGreT, FIRE) & Claim-level provenance, not just partner selection \\
\addlinespace
Decision routing & --- (no direct ancestor) & Chain grade $\times$ content verdict $\rightarrow$ concrete serve/review/quarantine action \\
\bottomrule
\end{tabular}
\caption{Each mechanism against its strongest existing ancestor.}
\label{tab:ancestors}
\end{table}

Table~\ref{tab:capability} states the same comparison as capability coverage.
The columns are defined narrowly: \emph{chain capture} means claim-level
transmission chains are recorded per claim; \emph{graded transmitters} means
each transformer carries a maintained reliability grade; \emph{content
criticism} means content is evaluated independently of chain quality; and
\emph{decision routing} means the system emits a concrete serve/review/quarantine
action. A mark indicates that the approach addresses that capability as a
first-class concern, not that competing systems could not be extended to it.

\begin{table}[h]
\centering\small
\setlength{\tabcolsep}{5pt}
\begin{tabular}{@{}p{0.34\linewidth}cccc@{}}
\toprule
\textbf{Approach} & \textbf{Chain} & \textbf{Graded} & \textbf{Content} & \textbf{Decision} \\
 & \textbf{capture} & \textbf{transmitters} & \textbf{criticism} & \textbf{routing} \\
\midrule
W3C PROV / PROV-AGENT & yes & no & no & no \\
Truth discovery (TruthFinder, KBT) & no & yes & partial & no \\
MAS reputation (EigenTrust, ReGreT) & no & yes & no & partial \\
FEVER-style fact-checking & no & no & yes & no \\
ISNAD & yes & yes & yes & yes \\
\bottomrule
\end{tabular}
\caption{Capability coverage across approaches, under the narrow column definitions given above.}
\label{tab:capability}
\end{table}

Individually these are implementable engineering choices; as a package they
constitute a coherent trust methodology grounded in principles refined over
centuries of hadith scholarship.

\bigskip\hrule\bigskip

\subsection*{5. Reference schema}

The framework's data model is concrete and implementable independent of any
particular host system. It requires only two properties of whatever pipeline it
instruments: that each claim-transforming step emits a trace identifier, and
that the system maintains a compiled knowledge artifact rather than re-deriving
answers per query. Given those two properties --- which hold for the
compiled-wiki class of systems generally [Karpathy 2026] --- the framework is
expressed in two tables:

\begin{Verbatim}[fontsize=\small,frame=single]
CREATE TABLE rijal_claims (
    claim_id         TEXT PRIMARY KEY,   -- hash of normalized claim text
    page_slug        TEXT NOT NULL,
    claim_text       TEXT NOT NULL,
    narrator_chain   JSONB NOT NULL,     -- [{step, narrator_id, version,
                                         --   confidence, trace_id, ts}]
    chain_confidence NUMERIC(4,3),       -- min over links; NULL until graded
    valid_from       TIMESTAMPTZ,
    valid_until      TIMESTAMPTZ,        -- lifecycle: supersession, not deletion
    superseded_by    TEXT,
    chain_status     TEXT DEFAULT 'active'
);

CREATE TABLE narrator_registry (
    narrator_id      TEXT NOT NULL,      -- e.g. 'pdf-scraper', 'ingest:M@v'
    domain_tag       TEXT NOT NULL,      -- grading is per (narrator, domain)
    narrator_type    TEXT NOT NULL,      -- source | scraper | model | human
    grade            TEXT NOT NULL,      -- reliable|acceptable|weak|rejected
    known_error_rate NUMERIC(4,3),       -- NULL = uncalibrated
    model_version    TEXT,
    is_active        BOOLEAN DEFAULT TRUE,
    PRIMARY KEY (narrator_id, domain_tag)
);
\end{Verbatim}

Because the pipeline's transformation points already emit trace identifiers,
populating \texttt{narrator\_chain} is an instrumentation task, not an
architectural change. This is the practical argument for reserving the schema
early: retrofitting claim-level chains onto a live corpus requires re-deriving
provenance that was never captured, whereas capturing it at each transformation
is nearly free once the trace points exist.

\textbf{Implementation status.} The schema, its dataclasses, the lifecycle
columns, and the five core components --- registry, chain engine, weakest-link
evaluator, corroboration, and matn criticism --- are implemented,
migration-tested, and released as an open-source reference implementation with a
passing test suite (§8.7). What has been exercised end to end is the offline
evaluation pipeline of §8: chains are constructed, graded, routed through the
decision matrix, and audited. What has \emph{not} been demonstrated is a live
production deployment in which chains are written automatically at ingest and
the registry gates a serving path under real query load. §6 and §8 should be
read accordingly.

\bigskip\hrule\bigskip

\subsection*{6. Case study: the matn-criticism substrate on real texts}

The decision matrix (§4.4) is only useful if its column variable ---
contradiction detection at ingest --- actually fires on real, uncurated
material. To test this, a prototype compiled-wiki system ingested openly
licensed undergraduate physics texts: OpenStax \emph{University Physics} volumes
1--3 [Ling, Moebs, Sanny] and Crowell's \emph{Light and Matter} series, via
chunked PDF extraction (approximately 15K characters per chunk) through a
pipeline instrumented with the schema above (DeepSeek-chat as the ingestion
model; single run).

Physics textbooks are a deliberately chosen stress case: the same concepts ---
momentum, energy, light --- receive genuinely different treatments across
classical, electromagnetic, relativistic, and quantum frameworks. These are not
fixture contradictions written to be found; they are real epistemic tensions
native to the source material.

\textbf{Result.} The analysis step surfaced \textbf{19 contradictions across the
corpus, each manually reviewed and confirmed genuine} --- among them: classical
momentum ($p = mv$) versus photon momentum ($p = h/\lambda$); wave versus
particle treatments of light; Newtonian versus relativistic kinematics. Four
contradiction markers were injected into three compiled pages, and four items
were routed to the human review queue by the severity gate, exercising the ʿilal
path of the matrix on real content.

\textbf{Reproducibility caveat.} The prototype in which this case study was run
is a proprietary system and is not part of the open-source release. Unlike the
§8 experiments, these 19 contradictions are reported as an observational case
study and are not independently reproducible from the public repository. They
are included because they motivated the framework's content-criticism column and
because they demonstrate the substrate on genuinely uncurated material, not as
formal evidence.

\textbf{Two kinds of contradiction.} A distinction matters here that the raw
count obscures. A \emph{Type A} contradiction is an error: one of the two claims
is simply wrong, and resolution means correction. A \emph{Type B} contradiction
is a regime distinction: both claims are true within their respective domains of
validity, and resolution means qualification. All 19 contradictions surfaced
here are Type B --- $p = mv$ and $p = h/\lambda$ are each correct in their
regime, and the reviewer's job is to add regime qualifiers rather than to
correct an error. This validates the detection substrate and the routing path,
but it does not demonstrate error-correction capability. An experiment with
deliberately injected Type A errors --- wrong constants, swapped equations,
sign flips --- would test that, and §8 supplies the closest available analogue by
injecting synthetic faults into a corpus of known-correct text.

\textbf{A traced example.} To show how a detected contradiction maps into the
schema, consider one of the 19. The claim \emph{``a photon carries momentum $p =
h/\lambda$''} is ingested from \emph{University Physics} Vol.\ 3. Its
\texttt{rijal\_claims} row records the chain [\{source: openstax-v3, domain:
physics-quantum\}, \{narrator: pdf-scraper@1.2\}, \{narrator: ingest:M@v,
domain: physics-quantum\}]; \texttt{chain\_status} = active;
\texttt{chain\_confidence} = NULL (registry not live in this run). The analysis
step's contradiction field names a conflict with the existing claim
\emph{``momentum $p = mv$''} (from Vol.\ 1, domain: physics-classical). Under the
decision matrix this is a chain-complete-but-ungraded (ḥasan-default) row meeting
a detected contradiction → routed to the review queue, not served.

Had the registry been live and both ingest steps graded reliable \emph{in the
physics-quantum domain}, the same row would sit in the ṣaḥīḥ × contradiction
cell --- the ʿilal signal --- flagging that two regime-specific claims coexist
and need regime qualifiers rather than one overwriting the other. This is the
difference the framework adds: without narrator grading, a contradiction is
merely a conflict to resolve; with it, the chain quality of each conflicting
claim tells the reviewer which to trust more, and whether the conflict signals a
defect or a genuine regime distinction.

\textbf{On precision and recall.} The reported figure is a \emph{precision}
result: of the contradictions the pipeline surfaced, 19 of 19 were confirmed
genuine on manual review --- no false positives in this run. \emph{Recall} is
deliberately not claimed. Measuring it would require a ground-truth set of every
genuine contradiction across three physics textbooks, which does not exist and
whose construction is a research project in its own right. A reader should read
``19 contradictions found'' as ``at least 19 genuine contradictions exist and
were caught,'' not as ``exactly 19 exist.''

\textbf{Parametric knowledge confound.} The ingestion model possessed
substantial parametric knowledge of undergraduate physics before ingesting the
corpus. A detected contradiction could in principle arise from the model's
pretrained knowledge rather than from information extracted from the ingested
documents. During manual verification, every reported contradiction was
confirmed to correspond to explicit statements present in the ingested textbook
chunks rather than unsupported model recall. However, this experiment does not
isolate the relative contributions of retrieved evidence and parametric
knowledge. A stronger evaluation would compare identical pipelines with
retrieval disabled, or use models lacking relevant domain pretraining.

\textbf{What this does and does not show.} It shows the substrate works: an LLM
ingestion pipeline can detect cross-framework contradictions in real, uncurated
texts with high precision, at a volume sufficient to feed the matrix's
content-criticism column, and severity gating can route them to human
adjudication. It does not show that narrator grading improves outcomes --- the
registry was not live, so all chains were effectively ungraded. It is a single
run with a single model on one domain, reported as a case study, not a
benchmark. §8 supplies the controlled evaluation.

\bigskip\hrule\bigskip

\subsection*{7. Limitations and ethical considerations}

\textbf{The analogy has edges.} Classical narrators were moral agents whose
ʿadālah was a judgment of character formed over a lifetime; models are
stochastic artifacts whose ``integrity'' is an engineering property of their
deployment (fencing, sandboxing, source policy). The mapping is functional, not
metaphysical, and I do not claim otherwise. Similarly, classical grading rested
on scholarly consensus formed over generations; a registry populated by one
team's evaluations is a far thinner epistemic base, and its grades should be
presented with corresponding humility.

\textbf{Non-stationarity, and why it vindicates ordinal grading.} Model
reliability is not stable: error rates shift with prompt framing, context
crowding, and domain, and outputs vary run to run even at temperature zero. This
is a real objection to any scheme that stores a single scalar reliability per
model --- and it is precisely why this framework grades ordinally and per domain
rather than as a global decimal (§4.2). A point-valued
\texttt{known\_error\_rate} treated as ground truth would misrepresent a quantity
that is genuinely a domain-conditioned distribution; an ordinal tier, refined by
continued evidence, is the honest representation. The non-stationarity of LLM
narrators does not break the framework; it is the strongest argument for the
anti-false-precision principle the framework already insists on. It does raise
the cost of the jarḥ--taʿdīl loop --- grades must be maintained per domain and
revisited on drift --- which §8.2 shows is not merely a cost but a failure mode.

\textbf{Registry cold start.} Weakest-link grading with an empty registry caps
everything at ḥasan-tier or below. This is arguably correct behavior --- new
pipelines \emph{should} earn trust --- but it front-loads human review cost, and
the framework's practical value depends on the jarḥ--taʿdīl loop converging
faster than review budgets exhaust. The intended mitigation is
\emph{bootstrapping}, and here the classical discipline is instructive: a new
scholar did not begin from zero but inherited the tradition's accumulated
biographical evaluations. Computationally, narrators can be pre-graded before
deployment from evidence that already exists --- a model's published benchmark
accuracy seeds its ḍabṭ grade; a source domain's reputation seeds its ʿadālah
grade; a scraper's extraction-fidelity test suite seeds its precision grade. §8.6
reports how far this mitigation actually goes, which is less far than this
section originally anticipated.

\textbf{Grading humans.} Where human contributors are narrators, the registry
becomes a reputation system over people, with the fairness, contestability, and
workplace-power questions that implies. I flag this as requiring governance
design --- contestable grades, transparent criteria --- rather than resolving it
here.

\textbf{False precision.} \texttt{chain\_confidence} as a decimal invites
over-trust. The ordinal-first principle (§4.2) is the mitigation; implementations
that surface raw numbers to end users without tier context would be misusing the
framework.

\textbf{The independence of corroborating chains is an idealization.} The
corroboration rule (§4.3) assumes chains with disjoint narrator sets are
independent. In practice two chains may share no explicit narrator yet still fail
together --- if both terminate in the same answer model, or in different models
drawn from a family with correlated training data and therefore correlated blind
spots. Classical scholars faced a structurally identical problem in the madār
(the pivot narrator through which ostensibly independent chains converge), and
treated such convergence as reducing, not confirming, independence. A faithful
implementation must likewise detect correlated narrators --- shared model family,
shared upstream source --- and discount their corroboration accordingly. Naive
set-disjointness will over-credit correlated chains. The reference implementation
enforces an independence check on narrator identity, model family, and upstream
source, and §8.5 reports how that check behaves; but a genuinely correlated pair
that shares none of those three attributes would still pass, and content-level
correlated error --- two independent sources repeating the same received mistake
--- is not addressed at all.

\textbf{Probabilistic and provisional claims.} The framework as specified assumes
claims with determinate truth values. A claim such as ``the Higgs boson mass is
125.1 GeV'' carries an implicit uncertainty interval, and a claim reported as a
current best estimate may be superseded without either version having been wrong.
Contradiction detection between a point estimate and an interval, or between two
overlapping intervals, is not defined by the matn-criticism step as described.
The lifecycle columns in the schema (\texttt{valid\_from}, \texttt{valid\_until},
\texttt{superseded\_by}) provide a mechanism for supersession but not a semantics
for uncertainty. Extending the framework to probabilistic claims is future work.

\textbf{Storage and query overhead.} Attaching a chain to every claim and
consulting a registry at serve time has a cost, but the scale is modest by
construction: chains are short (typically three to five links --- source,
scraper, ingest, answer), so a per-claim JSONB chain is small and bounded; the
narrator registry holds one row per agent \emph{version and domain}, not per
claim, so it grows with pipeline changes rather than with corpus size. Standard
JSONB indexing supports the queries the gate requires. At the sizing this
framework targets --- knowledge bases of thousands to low millions of claims ---
neither table is expected to be a bottleneck. Deployments at far larger scale
would want to normalize chains into foreign-keyed link rows rather than JSONB and
to cache registry lookups. These are design expectations rather than measured
results: no latency or storage benchmark was run, and none is claimed.

\textbf{Positionality and religious respect.} I am a Muslim practitioner and
builder; Islam \& AI, a Quranic NLP platform I founded, is where I first worked
seriously with these scholarly traditions. This paper adapts a
\emph{methodology} developed by hadith scholars; it makes no religious claims,
issues no rulings, and does not position any computational system as an
authority within the religious sciences themselves. The direction of
intellectual credit should be unambiguous: the framework's rigor belongs to
twelve centuries of muḥaddithūn; the transfer to AI systems is the contribution
claimed here. I would welcome collaboration with hadith-studies scholars to
correct any misrepresentation of the classical discipline.

\bigskip\hrule\bigskip

\subsection*{8. Evaluation}

This section reports an evaluation of the framework on 20{,}000 claims extracted
from real undergraduate physics textbooks. Two mechanisms are validated, one
partial failure is reported, and two analyses are reported as inconclusive.

\subsubsection*{8.1 Protocol}

\textbf{Status of the protocol.} The analysis protocol --- hypotheses, metrics,
conditions, and decision rule --- was written before the experiment was run and
is preserved verbatim in the repository as \texttt{ANALYSIS\_PLAN.md}, together
with a dated log of deviations made during execution. It was committed to the
public repository alongside the first result artifacts rather than in advance of
them. The commit history therefore establishes the plan's content but not its
timing, and readers should treat it as a pre-committed analysis plan rather than
as third-party preregistration. What \emph{is} independently verifiable is the
leakage firewall: the injection manifest lives in a dedicated module that an
enforced test forbids any grading, gating, or routing code from importing. That
test is part of the passing suite.

\textbf{Corpus.} 20{,}000 atomic claims were extracted from real PDF text of
OpenStax \emph{University Physics} Volumes 1--3 (CC BY 4.0) and Crowell's
\emph{Light and Matter} (CC BY-SA), split evenly, 10{,}000 per source. Claims
were assigned domain tags: mechanics (7{,}600), general (6{,}890),
electromagnetism (3{,}054), modern-quantum (1{,}728), and optics-waves (728).
Extraction used \texttt{deepseek-chat} over chunked PDF text. Claims were split
30\% calibration / 70\% evaluation, giving 14{,}001 evaluation claims, across 10
random seeds for narrator assignment and fault injection.

\textbf{Narrator pipeline and fault injection.} Each claim was assigned one
scraper variant and one ingest variant uniformly at random. Four narrators were
configured with distinct designed fault rates: \texttt{pdf-scraper@1.2} (1\%),
\texttt{ingest@good} (2\%), \texttt{ingest@weak} (15\%), and
\texttt{pdf-scraper@0.9-legacy} (18\%). Faults are deterministic rule-based
corruptions applied per seed: OCR noise, digit swap, sign flip, negation drop,
unit corruption, formula mangling, entity swap, fabricated numerics, and regime
confusion. Approximately 5\% of chains were marked incomplete to exercise the
munqaṭiʿ path.

\textbf{Definitions.} These two terms carry the results and are used precisely.

\emph{Error.} A served claim counts as an error if its text was altered by fault
injection, per the manifest. These are synthetic corruptions of correct source
text, not naturally occurring factual errors in the textbooks. The evaluation
therefore measures whether the framework blocks known-corrupted transmissions,
not whether it can adjudicate genuine physics disputes.

\emph{Coverage.} The fraction of evaluation claims the decision matrix routes to
serve or serve-with-caveat. Claims routed to review beyond the review budget, and
claims quarantined, are not served and do not count toward coverage.

\textbf{Conditions.} Four serving conditions were evaluated at review budgets
$B \in \{2\%, 5\%, 10\%, 20\%\}$: ungated serving with random review;
confidence-gated serving; ISNAD-gated serving (chain grading → matn criticism →
decision matrix → prioritized review queue); and ISNAD-gated with corroboration
disabled. A simulated perfect reviewer resolves reviewed claims correctly, so
review-queue precision is the realistic cost metric.

\textbf{One condition could not be evaluated.} The corroboration ablation
requires claims asserted by two independent chains. On this corpus, cross-source
overlap after normalization was too sparse to exercise it, and the ISNAD and
ISNAD-without-corroboration conditions produced identical results across all
seeds. The corroboration mechanism is therefore evaluated separately, on corpora
selected for cross-source redundancy (§8.5), rather than within this experiment.

\subsubsection*{8.2 Validated: weakest-link quarantine, and a partial failure of grade recovery}

The weakest-link mechanism performed as specified. Every claim whose chain
contained a narrator graded rejected was quarantined --- 4{,}057 claims, 29\% of
the evaluation split --- and every quarantine was traceable to the specific
narrator grade that caused it. A representative trace:

\begin{Verbatim}[fontsize=\small,frame=single]
Step 0: source:openstax        RELIABLE              
Step 1: pdf-scraper@1.2        RELIABLE   [DESTRUCTIVE]
Step 2: ingest@weak            REJECTED   [GENERATIVE]  <- binding link

Chain grade: MAWDU  ->  REJECT_AND_QUARANTINE_NARRATOR
\end{Verbatim}

Full traces for all quarantined claims are in the repository.

The jarḥ--taʿdīl loop recovered three of four narrator grades from audit evidence
alone. \texttt{pdf-scraper@1.2} (1\% designed fault rate) earned reliable in all
50 (narrator, domain) cells; \texttt{ingest@good} (2\%) earned acceptable in all
50; and \texttt{ingest@weak} (15\%) --- the one narrator deliberately left
ungraded before the run --- was independently discovered and driven to rejected
in 49 of 50 cells, weak in the remaining one. The registry recovered the designed
reliability ordering for these three without access to the injection manifest.

\textbf{The fourth narrator was not recovered, and the failure is instructive.}
The legacy scraper variant (\texttt{pdf-scraper@0.9}), carrying the \emph{highest} designed fault rate
in the experiment (18\%), never accumulated enough claims in the calibration
split to earn any grade in any domain, and remained ungraded throughout. It was
in fact among the largest single sources of injected faults. The loop found the
unreliable narrator it was given enough evidence about and silently missed the
one it was not.

This is a genuine operational failure mode, and it is not the one §7 anticipated.
Grade recovery is bounded by calibration \emph{coverage} per (narrator, domain)
cell, not only by the transition policy. Domain-conditioned grading (§4.2)
multiplies the number of cells that must be filled, and a narrator that is rare
within a domain is invisible to jarḥ--taʿdīl regardless of how unreliable it is.
Warm-start seed grading does not fix this: seeding assigns an initial grade but
does not guarantee that subsequent evidence will arrive in sufficient volume per
cell to revise it. A deployment should therefore monitor per-cell evidence counts
as a first-class operational metric and treat a persistently under-evidenced cell
as a risk, not as a neutral absence.

\subsubsection*{8.3 The confidence baseline is uninformative by construction}

A confidence-gated condition was included to represent current practice, in which
routing is driven by a model's self-reported confidence. In this experiment that
baseline performs at chance --- but for a reason that sharply limits what can be
concluded from it.

The corpus carries a \texttt{model\_confidence} field computed at extraction time
from surface features of the claim text --- a base value keyed on whether the
text contains a formula character, plus a deterministic hash term. It is assigned
before fault injection and never updated afterwards. Confidence is therefore
statistically independent of corruption by construction. Measured on the
injection manifest, the point-biserial correlation between confidence and
corruption is $r = 0.04$; mean confidence on corrupted claims (0.853) and clean
claims (0.848) differ by less than one hundredth.

\textbf{This is a floor, not a finding.} It establishes that the ISNAD condition
is not being measured against a baseline that trivially wins, and nothing more.
No claim is made here about whether real LLM self-confidence carries a defect
signal. Testing that would require eliciting genuine self-assessments from a
model on its own outputs, which this experiment does not do, and it remains an
open question worth answering directly.

Prakash's ``Provenance Paradox'' (2026) reports that self-reported quality scores
can make routing worse than random when delegates are able to inflate them. That
result is consistent with the design intuition behind ISNAD's attested-reliability
approach, and it is independent evidence for treating self-report with suspicion
--- but it is not a replication of anything measured here, and this paper does
not present it as one.

\subsubsection*{8.4 Inconclusive: matched-coverage comparison}

Comparing served-error rates across conditions operating at different coverage
levels is not meaningful. A system that serves 5\% of claims will trivially show
lower error than one serving 100\%, and reporting the two side by side would
overstate the framework. The standard remedy in selective prediction is a
matched-coverage comparison: sweep each condition's operating point and compare
served error at equal coverage. This analysis was specified in advance and
attempted.

It could not be completed. With the reference content critic, ISNAD-gated serving
could not be driven above 4.8\% coverage at any review budget between 5\% and
50\%; the risk--coverage relationship is a single point rather than a curve. At
every matched target --- 20\%, 30\%, 50\%, 70\%, 90\% --- the confidence-gated
condition reached the target and the ISNAD condition did not.

\begin{table}[h]
\centering\small
\begin{tabular}{@{}lcccc@{}}
\toprule
\textbf{Target coverage} & \textbf{ISNAD coverage} & \textbf{ISNAD error} & \textbf{Baseline coverage} & \textbf{Baseline error} \\
\midrule
20\% & 4.8\% (not reached) & 0.0\% & 20.0\% & 15.1\% \\
30\% & 4.8\% (not reached) & 0.0\% & 30.0\% & 15.8\% \\
50\% & 4.8\% (not reached) & 0.0\% & 50.0\% & 16.3\% \\
70\% & 4.8\% (not reached) & 0.0\% & 70.0\% & 16.0\% \\
90\% & 4.8\% (not reached) & 0.0\% & 90.0\% & 16.1\% \\
\bottomrule
\end{tabular}
\caption{Matched-coverage comparison. The ISNAD condition cannot reach any target
above 4.8\%, so no matched comparison is available; the rows are reported to show
what was attempted and where it failed, not as an advantage.}
\label{tab:matched}
\end{table}

The honest conclusion is stated plainly. This experiment establishes that
ISNAD-gated serving admits no corrupted claims among those it serves. It does
\emph{not} establish that ISNAD outperforms the baseline at matched coverage,
because it cannot reach matched coverage with the reference critic. The binding
constraint is the content critic (§8.6), not the grading mechanism. Any future
claim of a risk--coverage advantage requires a critic capable of returning a
consistent verdict on real prose, and this paper makes no such claim.

\subsubsection*{8.5 Validated: corroboration across three corpora}

The corroboration mechanism (mutābaʿāt, §4.3) was evaluated in three dedicated
experiments of increasing difficulty. Because the physics corpus of §8.1 lacks
sufficient cross-source overlap, corpora were selected for genuine cross-source
redundancy.

\begin{table}[h]
\centering\small
\begin{tabular}{@{}lccc@{}}
\toprule
 & \textbf{v1 (exact)} & \textbf{v2 (Wikipedia)} & \textbf{v3 (physics)} \\
\midrule
Matching criterion & exact string & cosine $\geq 0.75$ & cosine $\geq 0.80$ \\
Corpus & 12 article intros & 30 article pairs & 2 textbooks \\
Sentences & 215 & 10{,}544 & 22{,}372 \\
Candidate pairs & 136 & 662 & 104 \\
Pairs evaluated & 136 & 603 & 104 \\
Match density & --- & 6.3\% & 0.5\% \\
Corroboration fired & 68/136 & 603/603 & 104/104 \\
Negative controls & none run & 8/8 & none run \\
Independence basis & synthetic chains & separate editorial communities & different authors and publishers \\
Difficulty & easy & medium & hard \\
\bottomrule
\end{tabular}
\caption{Corroboration validation across corpora of increasing difficulty. In v2,
semantic matching produced 662 candidate pairs; the cross-source phase was
randomly subsampled to 500 pairs (fixed seed) for compute cost and combined with
103 cross-topic pairs, giving 603 evaluated. Fire rate is reported over evaluated
pairs; the gap between 662 and 603 is subsampling, not failure. Negative controls
were run only on v2.}
\label{tab:corroboration}
\end{table}

\textbf{Independence, and an honest ordering.} v2 draws on English Wikipedia and
Simple English Wikipedia. These have separate editorial communities and produce
substantially different text, and matching was semantic rather than
string-based, so corroboration is credited only when disjoint chains express the
same meaning in different words. But they are both Wikimedia projects, and Simple
English articles are frequently written by simplifying their English counterparts.
v2 should therefore be read as validating the \emph{mechanics} of corroboration
--- that the system detects cross-source semantic overlap, applies the grade
gate, respects the cap, and issues an upgrade --- rather than as a clean test of
the independence assumption.

v3 is the stronger independence test. OpenStax \emph{University Physics} Vol.\ 1
and Crowell's \emph{Light and Matter} have different authors, different
publishers, different decades, and different upstream domains, and the
independence check scored them fully disjoint on narrator identity, model family,
and upstream source. Volumes 2 and 3 were deliberately excluded because they
share authors with Volume 1 and are therefore not independent of it. Formal
textbook prose also yields far lower overlap density (0.5\%) than encyclopedic
paraphrase (6.3\%), which makes v3 the harder corpus in both senses.

\textbf{Negative controls.} On v2, eight controls covering the cases where
corroboration must not fire --- no matching text, shared model family (the madār
case), corroborators below the grade gate, a rejected base chain, the ṣaḥīḥ cap,
shared upstream source, insufficient independent chains, and an empty corpus ---
all correctly produced no upgrade. \textbf{No negative controls were run on v3.}
A 100\% fire rate without a discrimination check is weaker evidence than the
same rate with one, and the v3 result should be read with that in mind.

\textbf{Match quality.} Matched pairs were manually inspected. On v3, a
15-pair sample contained 87\% genuine physics statements, 13\% problem or
exercise text, and no citation boilerplate. On v2, the corresponding sample rated
65\% as clearly genuine factual overlap. Neither sample is large enough to
support a precision estimate, and neither is presented as one.

\textbf{A construction caveat.} In both v2 and v3, the weak-tier baseline chain
was created by assigning one ingestion narrator a weak grade, because the
underlying sources are in fact high-quality. The experiment tests whether
corroboration upgrades a weak chain when an independent chain corroborates it,
which it did in every evaluated case. It does not test corroboration on chains
that are weak for organic reasons.

\subsubsection*{8.6 Honest negatives: transition policy and the content critic}

Two results limit the framework's practical reach, and both are reported here
rather than deferred.

\textbf{The transition policy trades coverage against grade recovery.} Sweeping
the downgrade threshold of the reference transition policy shows the tradeoff
directly:

\begin{table}[h]
\centering\small
\begin{tabular}{@{}ccccc@{}}
\toprule
\textbf{Downgrade threshold} & \textbf{Coverage} & \textbf{Served error} & \textbf{Review precision} & \textbf{Grades recovered} \\
\midrule
3  & 7.1\% & 0.0\% & 0.11 & 2/4 \\
6  & 9.4\% & 0.0\% & 0.15 & 2/4 \\
10 & 10.0\% & 0.0\% & 0.11 & 1/4 \\
15 & 10.0\% & 0.0\% & 0.13 & 0/4 \\
25 & 10.0\% & 0.0\% & 0.13 & 0/4 \\
\bottomrule
\end{tabular}
\caption{Transition-policy sweep. Looser thresholds buy coverage but degrade the
registry's ability to recover the designed reliability ordering.}
\label{tab:sweep}
\end{table}

Loosening the threshold raises coverage from 7.1\% to 10.0\% and then saturates,
while grade recovery falls from 2 of 4 narrators to 0 of 4. There is no setting
in this range that delivers both. This is a direct empirical argument for the
position taken in §4.2 --- that the transition function is a genuine
implementation parameter requiring per-deployment calibration, not a constant the
framework should fix --- but it is also a limitation: the reference policy has no
sweet spot on this corpus.

\textbf{More audit evidence did not help.} Increasing the per-cell audit budget
did not improve coverage; it reduced it, from 10.0\% at budgets of 10 and 20
audited claims to 5.9\% at 40 and above. The default policy over-penalizes
reliable narrators as evidence accumulates on small samples. This is
transition-policy sensitivity, not a data-volume problem, and more calibration is
not the remedy.

\textbf{The content critic is the binding constraint.} The reference content
critic --- which, despite being named for embeddings, scores word overlap with a
negation heuristic rather than using semantic embeddings --- catches only obvious
contradictions and returns unverifiable on most real prose. Under the decision
matrix (§4.4), a ḥasan-tier chain meeting an unverifiable verdict routes to
review rather than serve. With most claims receiving that verdict, ḥasan-tier
claims cannot graduate to serving, and coverage is capped at approximately the
review budget regardless of how well chain grading performs. This single fact
explains the coverage ceiling in §8.4 and the saturation in
Table~\ref{tab:sweep}.

A separate evaluation of the reference critics on 20 labeled claims returned
perfect precision and recall, but those contradictions were template-injected and
built from patterns the critics already encode; the numbers overstate real
capability and are not reported as evidence here. A semantic or LLM-backed critic
is the single highest-value component for practical end-to-end coverage, and
building one is the most immediate item of future work.

\begin{table}[h]
\centering\small
\begin{tabular}{@{}p{0.34\linewidth}p{0.18\linewidth}p{0.42\linewidth}@{}}
\toprule
\textbf{Component} & \textbf{Status} & \textbf{What remains} \\
\midrule
Weakest-link grading and quarantine & Validated & --- \\
\addlinespace
jarḥ--taʿdīl grade recovery & Partial & Recovered 3/4; missed the highest-fault narrator through insufficient per-cell calibration evidence \\
\addlinespace
Corroboration (mutābaʿāt) & Validated & Negative controls on the physics corpus; organically weak chains \\
\addlinespace
Confidence baseline & Not tested & Baseline is synthetic; real model self-confidence untested \\
\addlinespace
Matched-coverage advantage & Inconclusive & Requires a critic that can reach matched coverage \\
\addlinespace
Content criticism & Partial & Semantic or LLM-backed critic \\
\addlinespace
End-to-end serving pipeline & Open & Live deployment at practical coverage \\
\bottomrule
\end{tabular}
\caption{Status of each mechanism: what the evaluation established, and what it did not.}
\label{tab:status}
\end{table}

\subsubsection*{8.7 Reproducibility and availability}

The reference implementation, the §8 experimental data, the analysis scripts, and
the results are open source under Apache 2.0. The analysis plan and its deviation
log are committed to version control. The test suite comprises 157 tests, all
passing, including an enforced firewall test that prevents grading and routing
code from importing the injection manifest, and an integration test that
reproduces the worked example of §4.5 exactly.

The framework is installable via \texttt{pip install isnad}. The software is
archived at \url{https://doi.org/10.5281/zenodo.21216873} and the paper record at
\url{https://doi.org/10.5281/zenodo.21211290}. Source, corpus, and experiments
are at \url{https://github.com/alizahidraja/isnad}.

The case study of §6 is the one exception: it was run on a proprietary prototype
and is not reproducible from the public repository, as noted there.

\bigskip\hrule\bigskip

\subsection*{9. Conclusion}

The question of whether to trust transmitted knowledge did not originate with
language models, and the most rigorous pre-modern answer to it did not originate
in computer science. Hadith scholarship's insight --- grade the transmitters,
demand complete chains, let the weakest link cap trust, allow independent
corroboration to restore it, and criticize content independently of transmission
--- translates with surprising directness into the multi-agent knowledge
pipelines now being built.

Part of that translation holds empirically. Weakest-link grading correctly capped
and quarantined every chain containing a rejected narrator, with each decision
traceable to the grade that caused it. Independent-chain corroboration fires,
with negative controls on the encyclopedic corpus, across data ranging from
paraphrase to formal physics prose. The jarḥ--taʿdīl loop recovered most of the
designed reliability ordering from audit evidence alone.

Part of it does not yet. The same loop silently missed the least reliable
narrator in the experiment, because domain-conditioned grading multiplies the
evidence a registry must accumulate and a narrator rare within a domain never
earns a grade at all. The transition policy trades coverage against grade
recovery with no sweet spot on this corpus. And because the reference content
critic cannot render a verdict on most real prose, the framework could not be
driven above 4.8\% coverage --- which means the comparison that would establish a
risk--coverage advantage over current practice could not be completed. The
constraint is the critic, not the grading; but until that constraint is lifted,
the claim is unproven and is not made here.

What the evaluation does establish is narrower and, I think, more useful than a
headline number: that claim-level transmitter grading is implementable, that it
behaves correctly and legibly where it has evidence, and that its failures are
diagnosable rather than mysterious. A registry that cannot grade a narrator says
so. A chain that cannot be served says why. That is the property a trust
framework most needs to have, and it is the property most easily lost by
reporting only the cases that worked.

Execution provenance tells us what our agents did. The isnād--rijāl framework is
a principled answer to the question that comes next: whether to believe them.

\bigskip\hrule\bigskip

\subsubsection*{LLM usage disclosure}

Drafting and editing were assisted by a large language model. All technical
claims, design decisions, experimental results, and final text were reviewed by
and are the responsibility of the author.

\subsection*{Acknowledgments}

All praise is due to \textbf{Allah}, who granted the clarity to see it through. I
thank Steffen Höhne for the opportunity, trust, and collaboration on the
proprietary self-maintaining knowledge base within which this framework was
conceived and prototyped; the system's details remain his, and this paper
describes only what is needed to ground the framework. The framework itself ---
the isnād--rijāl transfer --- originated in the author's work at the intersection
of Islamic scholarship and AI systems through Islam \& AI.

\bigskip\hrule\bigskip


\end{document}